\documentclass[11pt,a4paper]{article}
\usepackage{times,latexsym}
\usepackage{url}
\usepackage[T1]{fontenc}
\usepackage{amsmath}
\usepackage{graphicx}
\usepackage{xcolor}
\usepackage{subcaption}
\usepackage{grffile}
\usepackage{bbm}
\usepackage{enumitem}
\usepackage{booktabs}
\usepackage{multirow}
\usepackage{placeins}

\usepackage[hyperref]{acl2019}

\usepackage{layouts}

\definecolor{burgundy}{rgb}{0.5, 0.0, 0.13}
\definecolor{darkpastelgreen}{rgb}{0.01, 0.75, 0.24}
\definecolor{ferngreen}{rgb}{0.31, 0.47, 0.26}

\aclfinalcopy 
\def\aclpaperid{1338} 

\usepackage{todonotes}

\newcommand{\ignore}[1]{}%

\newcounter{mwc}

\newcounter{bd}

\newcounter{dd}

\newcounter{mfar}

\newcounter{wc}

\title{Handling Divergent Reference Texts when Evaluating\\Table-to-Text Generation}

\author{Bhuwan Dhingra$^\dag$
\thanks{\enskip Work done during an internship at Google.}
\qquad Manaal Faruqui$^\ddag$
\qquad Ankur Parikh$^\ddag$
\qquad Ming-Wei Chang$^\ddag$\\
{\bf Dipanjan Das$^\ddag$
\qquad William W. Cohen$^{\dag \ddag}$}\\
$^\dag$ Carnegie Mellon University\\
$^\ddag$ Google Research\\
\texttt{bdhingra@cs.cmu.edu}\\
\texttt{\{mfaruqui,aparikh,mingweichang,dipanjand,wcohen\}@google.com}
}

\date{}

\begin{document}
\maketitle
\begin{abstract}
Automatically constructed datasets for 
generating text from semi-structured data (tables),
such as WikiBio \citep{lebret2016neural},
often contain reference texts that
diverge from the information in the corresponding semi-structured data.
We show that metrics which rely solely on the reference texts,
such as BLEU and ROUGE,
show poor correlation with human judgments when those references diverge.
We propose a new metric, PARENT, which 
aligns n-grams from the reference and generated texts
to the semi-structured data before computing their precision and recall.
Through a large scale human evaluation study of
table-to-text models for WikiBio,
we show that PARENT correlates with human judgments better
than existing text generation metrics.
We also adapt and evaluate the information extraction based
evaluation proposed in \citet{wiseman2017challenges},
and show that PARENT has
comparable correlation to it,
while being easier to use. 
We show that PARENT is also applicable
when the reference texts are elicited from humans
using the data from the
WebNLG challenge.%
\footnote{Code and Data: \url{http://www.cs.cmu.edu/~bdhingra/pages/parent.html}}
\end{abstract}

\section{Introduction}


The task of generating natural language descriptions of structured data (such as tables) \cite{Kukich:1983:DKR:981311.981340,McKeown:1985:TGU:4047,Reiter:1997:BAN:974487.974490} has seen a growth in interest with the rise of sequence to sequence models that provide an easy way of encoding tables and generating text from them \citep{lebret2016neural,wiseman2017challenges,e2e-nlg,gardent2017creating}.

\begin{figure*}[!tb]
    \centering
    \includegraphics[width=\textwidth]{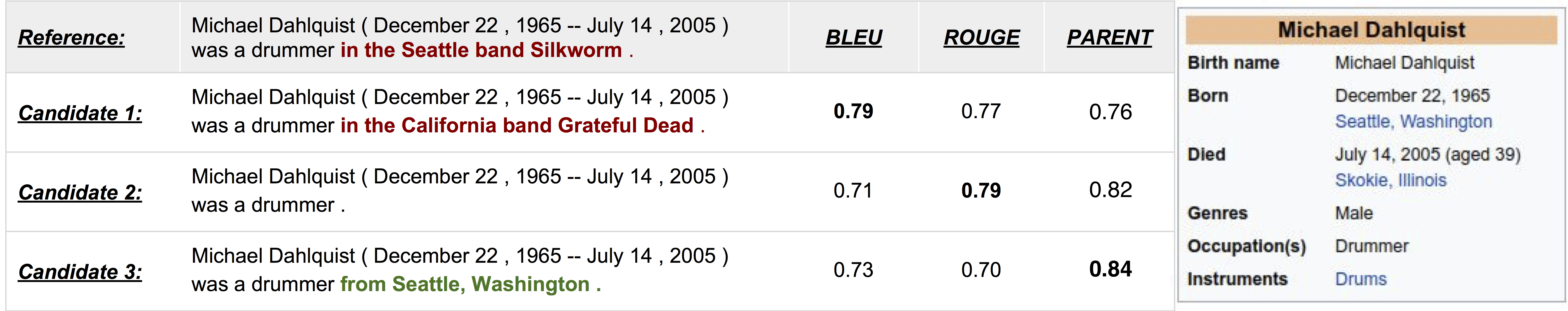}
    \caption{A table from the WikiBio dataset (right), its reference description and three hypothetical generated texts with scores assigned to them by automatic evaluation metrics. Text which cannot be inferred from the table is in {\color{burgundy}{\textbf{red}}}, and text which can be inferred but isn't present in the reference is in {\color{ferngreen}{\textbf{green}}}. PARENT is our proposed metric.}
    \label{fig:divergence-example}
\end{figure*}

For text generation tasks, the only gold standard metric is to show the output to humans for judging its quality, but this is too expensive to apply 
repeatedly anytime small modifications are made to a system. 
Hence, automatic metrics that compare the generated text to one or more reference texts are routinely used to compare models \citep{bangalore2000evaluation}. 
For table-to-text generation, automatic evaluation has largely relied on BLEU \citep{bleu} and ROUGE \citep{rouge}. 
The underlying assumption behind these metrics is that the reference text is \textit{gold-standard}, i.e., it is the ideal target text that a system should generate. In practice, however, when datasets are collected automatically and heuristically, the reference texts are often not ideal. Figure~\ref{fig:divergence-example} shows an example from the WikiBio dataset \citep{lebret2016neural}. Here the reference contains extra information which no system can be expected to produce given only the associated table.
We call such reference texts \textit{divergent} from the table. 

We show that existing automatic metrics, including BLEU, correlate poorly with human judgments when the evaluation sets contain divergent references (\S \ref{sec:results}).
For many table-to-text generation tasks,
the tables themselves are in a pseudo-natural language format
(e.g., WikiBio, WebNLG \citep{gardent2017creating}, and E2E-NLG \citep{dusek2019e2e}).
In such cases we propose to compare the generated text to the
underlying table as well to improve evaluation. We
develop a new metric, PARENT (\textbf{P}recision \textbf{A}nd \textbf{R}ecall of \textbf{E}ntailed \textbf{N}-grams from the \textbf{T}able) (\S \ref{sec:parent}).
When computing precision, PARENT effectively uses a union of the reference
and the table, to reward correct information missing from the
reference.
When computing recall, it uses an intersection of the reference
and the table, to ignore extra incorrect information in the reference.
The union and intersection are computed with the help of
an entailment model to decide if a text n-gram
is entailed by the table.%
\footnote{Here ``entailed'' means can be reasonably
inferred from the corresponding table.
In practice, we use simple lexical entailment models
to determine this.
}
We show that this method is more effective than using the table
as an additional reference.
Our main contributions are:
\begin{itemize}[leftmargin=*]
\setlength\itemsep{0em}
\item We conduct a large-scale human evaluation of the outputs from $16$ table-to-text models
on $1100$ examples from the WikiBio dataset, many of which have divergent references (\S \ref{sec:human}).
\item We propose a new metric, PARENT (\S \ref{sec:parent}),
and show that it improves correlation with human judgments over
existing metrics, both when comparing similar systems
(such as different hyperparameters of a neural network)
and when comparing vastly different systems
(such as template-based and neural models).%
\item We also develop information extraction based metrics, inspired from
\citet{wiseman2017challenges}, by training a model to extract
tables from the reference texts (\S \ref{sec:extraction}).
We find that these metrics have comparable
correlation to PARENT, with the latter being easier to use out of the box.
\item We analyze the sensitivity of the metrics to divergence by collecting
labels for which references contain only information also present in the tables.
We show that PARENT maintains high correlation as the number of such examples is varied. (\S \ref{sec:noise}).
\item We also demonstrate the applicability of PARENT on the data released as part of the WebNLG challenge \citep{gardent2017creating}, where the references are elicited from humans,
and hence are of high quality (\S \ref{sec:results}). 
\end{itemize}

\section{Table-to-Text Generation}
We briefly review the task of generating natural language descriptions of semi-structured data, which we refer to as \textit{tables} henceforth \citep{Barzilay:2005:CCS:1220575.1220617,liang2009learning}. Tables can be expressed as
set of records $T = \{r_k\}_{k=1}^K$, where each record is a tuple \textit{(entity, attribute, value)}.
When all the records are about the same entity, we can truncate the records to \textit{(attribute, value)} pairs. For example, for the table in Figure~\ref{fig:divergence-example}, the records are \textit{\{(Birth Name, Michael Dahlquist), (Born, December 22 1965), ...\}}. The task is to generate a text $G$ which summarizes the records in a fluent and grammatical manner.\footnote{In some cases the system is expected to summarize all the records (e.g. WebNLG); in others the system is expected to only summarize the salient records (e.g. WikiBio).} For training and evaluation we further assume that we have a reference description $R$ available for each table.
We let $\mathcal{D}_M = \{(T^i, R^i, G^i)\}_{i=1}^N$ denote an evaluation set of tables, references and texts generated from a model $M$, and $R_n^i$, $G_n^i$ denote the collection of n-grams of order $n$ in $R^i$ and $G^i$, respectively.
We use $\#_{R_n^i}(g)$ to denote the count of n-gram $g$ in $R_n^i$,
and $\#_{G_n^i,R_n^i}(g)$ to denote the minimum of its counts in $R_n^i$ and $G_n^i$.
Our goal is to assign a score to the model, which correlates highly
with human judgments of the quality of that model.

\paragraph{Divergent References.}
In this paper we are interested in the case where
reference texts \textit{diverge} from the tables.
In Figure~\ref{fig:divergence-example}, the reference, though technically correct and fluent, mentions information
which cannot be gleaned from the associated table. It also fails to mention useful information which a generation system might correctly include (e.g. candidate $3$ in the figure). We call such references \textit{divergent} from the associated table.
This phenomenon is quite common -- in WikiBio we found that $62\%$ of the references mention extra information (\S \ref{sec:noise}). 
Divergence is common in human-curated translation datasets as well \citep{carpuat2017detecting,vyas2018identifying}.

How does divergence affect automatic evaluation? As a motivating example, consider the three candidate generations shown in Figure~\ref{fig:divergence-example}. Clearly, candidate $1$ is the worst since it ``hallucinates'' false information, and candidate $3$ is the best since it is correct and mentions more information than candidate $2$. However, BLEU and ROUGE, which only compare the candidates to the reference, penalize candidate $3$ for both excluding the divergent information in the reference (in red) and including correct information from the table (in green).\footnote{BLEU is usually computed at the corpus-level, however here we show its value for a single sentence purely for illustration purposes. The remaining BLEU scores in this paper are all at the corpus-level.} PARENT, which compares to both the table and reference, correctly ranks the three candidates.

\section{PARENT}
\label{sec:parent}

PARENT evaluates each instance $(T^i, R^i, G^i)$ separately,
by computing the precision and recall of $G^i$ against both $T^i$ and $R^i$.

\paragraph{Entailment Probability.}
The table is in a semi-structured form, and hence not directly comparable to the unstructured generated or reference texts. 
To bridge this gap, we introduce the notion of entailment probability,
which we define as the probability that the presence of an n-gram $g$
in a text is ``correct'' given the associated table.
We denote this probability as $w(g) = Pr(g \Leftarrow T^i)$.
Estimating this probability is in itself a challenging language understanding task, since the information in the table may be expressed in varied forms in text. 
Here, we describe two simple models of lexical entailment,
inspired by work on the Recognizing Textual Entailment Challenge \citep{rte}.
We found these simple models to be effective; while more sophisticated
models may be used if there are complex inferences between the table
and text,
they are beyond the scope of this paper.

\begin{enumerate}[leftmargin=*]
    \item \textbf{Word Overlap Model:} Let $\bar{T}^i$ denote all the lexical items present in the table $T^i$,
    including both attribute names and their values. Then,
        $w(g) = \sum_{j=1}^n \mathbbm{1}(g_j \in \bar{T}^i)/n$, 
    where $n$ is the length of $g$, and $g_j$ is the $j$th token in $g$.

    \item \textbf{Co-occurrence Model:} \citep{glickman2005probabilistic}
    Originally proposed for the RTE task, this model computes the 
    probability of a term $g_j$ in the n-gram being entailed by the table
    as the maximum of its probabilities of being entailed by each
    lexical item $v$ in the table:
    \begin{equation}
        Pr(g_j \Leftarrow T^i) = \max_{v \in \bar{T}^i} Pr(g_j \Leftarrow v).
    \end{equation}
    $Pr(g_j \Leftarrow v)$
    is estimated using
    co-occurrence counts from a training set of table-reference pairs.
    Then the overall probability of the n-gram being entailed is taken as
    the geometric average $w(g) = \left(\prod_{j=1}^n Pr(g_j \Leftarrow T^i)\right)^{1/n}$.%
    \footnote{\citet{glickman2005probabilistic} used a product instead of
    geometric mean. Here we use a geometric mean to ensure that n-grams of 
    different lengths have comparable probabilities of being entailed.}
\end{enumerate}
We note that these models are not sensitive to paraphrases between the table and text.
For tasks where this is important, embedding-based similarities may be used,
but those are beyond the scope of this paper.
Next we discuss how to compute the precision and recall of the generation.

\paragraph{Entailed Precision.}  When computing precision, we want to check what fraction of the n-grams in $G_n^i$ are correct. We consider an n-gram $g$ to be correct either if it occurs in the reference $R_n^i$\footnote{It is unlikely that an automated system produces the same extra n-gram as present in the reference, thus a match with the reference n-gram is considered positive. For example, in Figure~\ref{fig:divergence-example}, it is highly unlikely that a system would produce ``Silkworm'' when it is not present in the table.}, or if it has a high probability of being entailed by the table (i.e. $w(g)$ is high). Let $Pr(g \in R_n^i) = \frac{\#_{G_n^i,R_n^i}(g)}{\#_{G_n^i}(g)}$ denote the probability that an n-gram in $G_n^i$ also appears in $R_n^i$. Then, the entailed precision $E_p^n$ for n-grams of order $n$ is given by:
\begin{align}
    &E_p^n = \nonumber\\
    &\frac{\sum_{g \in G_n^i} \left[Pr(g \in R_n^i) + Pr(g \notin R_n^i) w(g)\right]\#_{G_n^i}(g)}{\sum_{g \in G_n^i}\#_{G_n^i}(g)}, \nonumber\\
    &= \frac{\sum_{g \in G_n^i} \#_{G_n^i}(g) w(g) + \#_{G_n^i,R_n^i}(g) [1 - w(g)]}{\sum_{g \in G_n^i}\#_{G_n^i}(g)}.
    \label{eq:precision}
\end{align}
In words, an n-gram receives a reward of $1$ if it appears in the reference, with probability $Pr(g \in R_n^i)$, and otherwise it receives a reward of $w(g)$. Both numerator and denominator are weighted by the count of the n-gram in $G_n^i$. 
$Pr(g \in R_n^i)$ rewards an n-gram for appearing as many times as it appears in the reference, not more.
We combine precisions
for n-gram orders $1$-$4$ using a geometric average, similar to BLEU:
\begin{align}
    E_p = \exp\left(\sum_{n=1}^4 \frac{1}{4} \log E_p^n\right)
\end{align}

\paragraph{Entailed Recall.} We compute recall against both the reference ($E_r(R^i)$), to ensure proper sentence structure in the generated text, and the table ($E_r(T^i)$), to ensure that texts which mention more information from the table get higher scores (e.g. candidate $3$ in Figure~\ref{fig:divergence-example}).
These are combined using a geometric average:
\begin{equation}
    E_r = E_r(R^i)^{(1-\lambda)} E_r(T^i)^\lambda
\label{eq:lambda}
\end{equation}
The parameter $\lambda$ trades-off how much the generated text should match the reference, versus how much it should cover information from the table.
The geometric average, which acts as an AND operation, ensures that the overall recall is high only when both the components are high. We found this necessary to assign low scores to bad systems which, for example, copy values from the table without phrasing them in natural language.

When computing $E_r(R^i)$, divergent references will have n-grams with low $w(g)$. We want to exclude these from the computation of recall, and hence their contributions are weighted by $w(g)$:
\begin{align}
    E_r^n(R^i) = \frac{\sum_{g \in R_n^i} \#_{G_n^i,R_n^i}(g)w(g) }{\sum_{g \in R_n^i} \#_{R_n^i}(g) w(g)}.
    \label{eq:recall}
\end{align}
Similar to precision, we combine recalls for $n=1$-$4$ using a geometric
average to get $E_r(R^i)$.

For computing $E_r(T^i)$, note that a table is a set of records $T^i = \{r_k\}_{k=1}^K$. For a record $r_k$, let $\bar{r}_k$ denote its string \textit{value} 
(such as \textit{``Michael Dahlquist''} or \textit{``December 22 1965''}). Then:
\begin{equation}
    E_r(T^i) = \frac{1}{K} \sum_{k=1}^K \frac{1}{|\bar{r}_k|} LCS(\bar{r}_k, G^i),
    \label{eq:table}
\end{equation}
where $\bar{r}_k$ denotes the number of tokens in the value string, and $LCS(x, y)$ is the length of the longest common subsequence between $x$ and $y$. 
The $LCS$ function, borrowed from ROUGE, ensures that 
entity names in $\bar{r}_k$ appear in the same order in the text as the table. Higher values of $E_r(T^i)$ denote that more records are likely to be mentioned in $G^i$.

The entailed precision and recall are combined into an F-score to give the PARENT metric for one instance.
The system-level PARENT score for a model $M$ is the average of instance level PARENT scores across the evaluation set:
\begin{equation}
    PARENT(M) = \frac{1}{N}\sum_{i=1}^N PARENT(G^i, R^i, T^i)
\end{equation}

\paragraph{Smoothing \& Multiple References.} The danger with geometric averages is that if any of the components being averaged become $0$, the average will also be $0$. Hence, we adopt a smoothing technique from \citet{chen2014systematic} that assigns a small positive value $\epsilon$ to any of $E_p^n$, $E_r^n(R^i)$ and $E_r(T^i)$ which are $0$. 
When multiple references are available for a table, we compute PARENT against each reference and take the maximum as its overall score, similar to METEOR \citep{meteor}.

\paragraph{Choosing $\lambda$ and $\epsilon$.}
To set the value of $\lambda$ we can tune it to maximize the
correlation of the metric with human judgments, when such data
is available. When such data is not available,
we can use the recall of the \textit{reference} against
the table, using Eq.~\ref{eq:table}, as the value of $1 -\lambda$.
The intuition here is that if the recall of the reference against
the table is high, 
it already covers most of the information, and
we can assign it a high weight in Eq.~\ref{eq:lambda}.
This leads to a separate value of $\lambda$ automatically
set for each instance.%
\footnote{For WikiBio, on average $\lambda=0.6$ using this heuristic.}
$\epsilon$ is set to $10^{-5}$ for all experiments.

\section{Evaluation via Information Extraction}
\label{sec:extraction}
\citet{wiseman2017challenges} proposed to use an auxiliary model, trained to extract structured records from text, for evaluation.
However, the extraction model presented in that work
is limited to the closed-domain setting of basketball game
tables and summaries.
In particular, they assume that each table has exactly the
same set of attributes for each entity,
and that the entities can
be identified in the text via string matching.
These assumptions are not valid for the open-domain
WikiBio dataset, and hence we train our own extraction
model to replicate their evaluation scheme.


Our extraction system is a pointer-generator network \citep{see2017pointer},
which learns to produce a linearized version of the table from the text.%
\footnote{
All \textit{(attribute, value)} pairs are merged into $1$ long string using
special separator tokens between them.}
The network learns which attributes need to be populated
in the output table, along with their values.
It is trained on the training set of WikiBio.
At test time we parsed the output strings into a set of
\textit{(attribute, value)} tuples and
compare it to the ground truth table.
The F-score of this text-to-table system was $35.1\%$,
which is comparable to other challenging open-domain settings
\citep{huang2017improving}.
More details are included in the Appendix~\ref{app:ie-system}.

Given this information extraction system,
we consider the following metrics for evaluation,
along the lines of
\citet{wiseman2017challenges}.
    \textbf{Content Selection (CS):} F-score for the
    \textit{(attribute, value)} pairs extracted from the
    generated text compared to those extracted from the
    reference.
    \textbf{Relation Generation (RG):} Precision for the
    \textit{(attribute, value)} pairs extracted from the
    generated text compared to those in the 
    ground truth table.
    \textbf{RG-F:} Since our task emphasizes the recall of information from
    the table as well, we consider another variant which computes the
    F-score of the extracted pairs to those in the table.
We omit the content ordering metric,
since our extraction system does not align
records to the input text.

\section{Experiments \& Results}
\label{sec:experiments}
In this section we compare several automatic
evaluation metrics by checking their correlation with the scores
assigned by humans to table-to-text models.
Specifically, given $l$ models $M_1, \ldots, M_l$, and their outputs on an evaluation set, we show these generated texts to humans to judge their quality, and obtain aggregated human evaluation scores for all the models, $\bar{h} = (h_1, \ldots, h_l)$ (\S \ref{sec:human}). Next, to evaluate an automatic metric, we compute the scores it assigns to each model, $\bar{a} = (a_1, \ldots, a_l)$, and check the Pearson correlation between $\bar{h}$ and $\bar{a}$ \citep{graham2014testing}.%
\footnote{We observed similar trends for Spearman correlation.}

\subsection{Data \& Models}
\label{sec:models}
Our main experiments are on the WikiBio dataset
\citep{lebret2016neural},
which is automatically constructed and contains many
divergent references.
In \S \ref{sec:webnlg} we also present results on the
data released as part of the WebNLG challenge.

%
\begin{table}[!tb]
\setlength\tabcolsep{1.2pt}
\small
\centering
\begin{tabular}{@{}ccccc@{}}
\toprule
\textbf{Name}    & \textbf{\begin{tabular}[c]{@{}c@{}}Beam\\ Size\end{tabular}} & \textbf{\begin{tabular}[c]{@{}c@{}}Length\\ Penalty\end{tabular}} & \textbf{\begin{tabular}[c]{@{}c@{}}Beam\\ Rescoring\end{tabular}} & \textbf{\begin{tabular}[c]{@{}c@{}}Human\\ Eval\end{tabular}} \\ \midrule
References      & --            & --                & --                    & 0.20 $\pm$ 0.03 \\ \midrule
Template         & --              & --                          & --             & -0.19 $\pm$ 0.04          \\
Seq2Seq          & 1             & 0       & No               & -0.28 $\pm$ 0.03          \\
Seq2Seq + Att & 1                 & 0       & No        & -0.12 $\pm$ 0.03          \\
PG-Net           & 1,4,8       & 0,1,2,3      & No,Yes   & 0.40 $\pm$ 0.03  \\ \bottomrule
\end{tabular}
\caption{Models used for WikiBio, with the human evaluation scores for these model outputs and the reference texts. PG-Net: Pointer-Generator network. 
Human scores computed using Thurstone's method \citep{tsukida2011analyze}.
}
\label{tab:models}
\end{table}

We developed several models of varying quality for generating text from the tables in WikiBio. This gives us a diverse set of outputs to evaluate the automatic metrics on. Table~\ref{tab:models} lists the models along with their hyperparameter settings and their scores from the human evaluation (\S\ref{sec:human}). Our focus is primarily on neural sequence-to-sequence methods since these are most widely used, but we also include a template-based baseline. All neural models were trained on the WikiBio training set. 
Training details and sample outputs are included in Appendices~\ref{app:hyperparams} \& \ref{app:outputs}.

We divide these models into two categories and measure correlation separately for both the categories. The first category, \textbf{WikiBio-Systems}, includes one model each from the four families listed in Table~\ref{tab:models}.
This category tests whether a metric can be used to compare different model families with a large variation in the quality of their outputs. 
The second category, \textbf{WikiBio-Hyperparams}, includes $13$ different hyperparameter settings of PG-Net \citep{see2017pointer},
which was the best performing system overall.
$9$ of these were obtained by varying the beam size and length normalization penalty of the decoder network \citep{45610}, and the remaining $4$ were obtained by re-scoring beams of size $8$ with the information extraction model described in \S \ref{sec:extraction}.
All the models in this category produce high quality fluent texts, and differ primarily on the quantity and accuracy of the information they express. Here we are testing whether a metric can be used to compare similar systems with a small variation in performance. This is an important use-case as metrics are often used to tune hyperparameters of a model.



\subsection{Human Evaluation}
\label{sec:human}

\begin{figure*}
    \centering
    \fbox{\includegraphics[width=\textwidth,trim={0 0.3cm 0 2.8cm},clip]{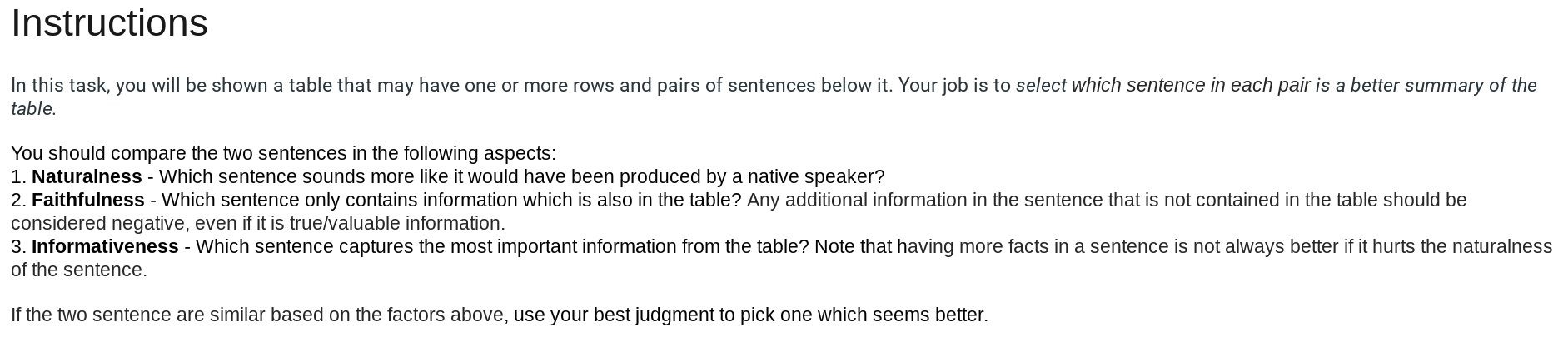}}
    \caption{Instructions to crowd-workers for comparing two generated texts.}
    \label{fig:instructions}
\end{figure*}

We collected human judgments on the quality of the $16$ models trained for WikiBio, plus the reference texts. Workers on a crowd-sourcing platform,
proficient in English,
were shown a table with pairs of generated texts, or a generated text and the reference, and asked to select the one they prefer.
Figure~\ref{fig:instructions} shows the instructions they were given.
Paired comparisons have been shown to be superior to
rating scales for comparing generated texts
\cite{callison2007meta}.
However, for measuring correlation the comparisons need to be aggregated into real-valued scores, $\bar{h} = (h_1, \ldots, h_l)$, for each of the $l=16$ models. For this, we use Thurstone's method \citep{tsukida2011analyze}, which assigns a score to each model based on how many times it was preferred over an alternative.

The data collection was performed separately for models in the \textbf{WikiBio-Systems} and \textbf{WikiBio-Hyperparams} categories. $1100$ tables were sampled from the development set, and for each table we got $8$ different sentence pairs annotated across the two categories, resulting in a total of $8800$ pairwise comparisons. Each pair was judged by one worker only which means there may be noise at the instance-level, but the aggregated system-level scores had low variance (cf. Table~\ref{tab:models}).
In total around $500$ different workers were involved in the annotation.
References were also included in the evaluation, and they received a lower score than PG-Net,
highlighting the divergence in WikiBio.

\subsection{Compared Metrics}
\textbf{Text only:} We compare
BLEU \citep{bleu},
ROUGE \citep{rouge},
METEOR \citep{meteor},
CIDEr and CIDEr-D \citep{cider}
using their publicly available implementations.

\noindent\textbf{Information Extraction based:} We compare the CS, RG and RG-F metrics discussed in
\S \ref{sec:extraction}.

\noindent\textbf{Text \& Table:}
We compare a variant of BLEU, denoted as
BLEU-T, where the values from the table
are used as additional references.
BLEU-T draws inspiration from iBLEU \citep{sun2012joint} but
instead rewards n-grams which match the table rather than
penalizing them.
For PARENT, we compare
both the word-overlap model (PARENT-W)
and the co-occurrence model (PARENT-C) for determining entailment.
We also compare versions where a single $\lambda$ is tuned
on the entire dataset to maximize correlation with human judgments, denoted as PARENT*-W/C.

\subsection{Correlation Comparison}
\label{sec:results}

\begin{table}[t]
\centering
\small
\begin{tabular}{@{}lccc@{}}
\toprule
\multirow{2}{*}{\textbf{Metric}} 
                                 & \textbf{\begin{tabular}[c]{@{}c@{}}WikiBio\\ Systems\end{tabular}} & \textbf{\begin{tabular}[c]{@{}c@{}}WikiBio\\ Hyperparams\end{tabular}} & \textbf{Average} \\ \midrule
ROUGE                            & 0.518$\pm$0.07$^{C,W}$      & -0.585$\pm$0.15$^{C,W}$                                                     & -0.034           \\
CIDEr                           & 0.674$\pm$0.06$^{C,W}$     & -0.516$\pm$0.15$^{C,W}$    & 0.079 \\
CIDEr-D                         & 0.646$\pm$0.06$^{C,W}$     & -0.372$\pm$0.16$^{C,W}$    & 0.137 \\
METEOR                           & 0.697$\pm$0.06$^{C,W}$      & -0.079$\pm$0.24$^{C,W}$                                                     & 0.309            \\
BLEU                             & 0.548$\pm$0.07$^{C,W}$      & 0.407$\pm$0.15$^{C,W}$                                                      & 0.478            \\ \midrule
CS                     & 0.735$\pm$0.06$^W$                & -0.604$\pm$0.16$^{C,W}$                                                     & 0.066            \\
BLEU-T                           & 0.688$\pm$0.11$^W$      & 0.587$\pm$0.14$^{C,W}$                                                     & 0.638           \\
RG                     & 0.645$\pm$0.07$^{C,W}$                & 0.749$\pm$0.12                                                      & 0.697         \\
RG-F                     & 0.753$\pm$0.06$^W$                & 0.763$\pm$0.12                                                      & 0.758         \\ \midrule
PARENT-C                      & 0.776$\pm$0.05$^W$      & 0.755$\pm$0.12                                                      & 0.766            \\
PARENT-W                      & 0.912$\pm$0.03      & 0.763$\pm$0.12                                                      & 0.838            \\ \midrule
\textit{PARENT*-C}                      & 0.976$\pm$0.01     & 0.793$\pm$0.11                                              & 0.885            \\
\textit{PARENT*-W}                      & \textbf{0.982}$\pm$\textbf{0.01}      & \textbf{0.844}$\pm$\textbf{0.10}         & \textbf{0.913}            \\ \bottomrule
\end{tabular}
\caption{Correlation of metrics with human judgments on WikiBio. 
A superscript of $C$/$W$ indicates that the correlation is significantly lower than that
of PARENT-C/W using a bootstrap confidence test for $\alpha=0.1$.}
\label{tab:results}
\end{table}

\begin{figure}[!tb]
    \centering
        \includegraphics[width=0.49\linewidth]{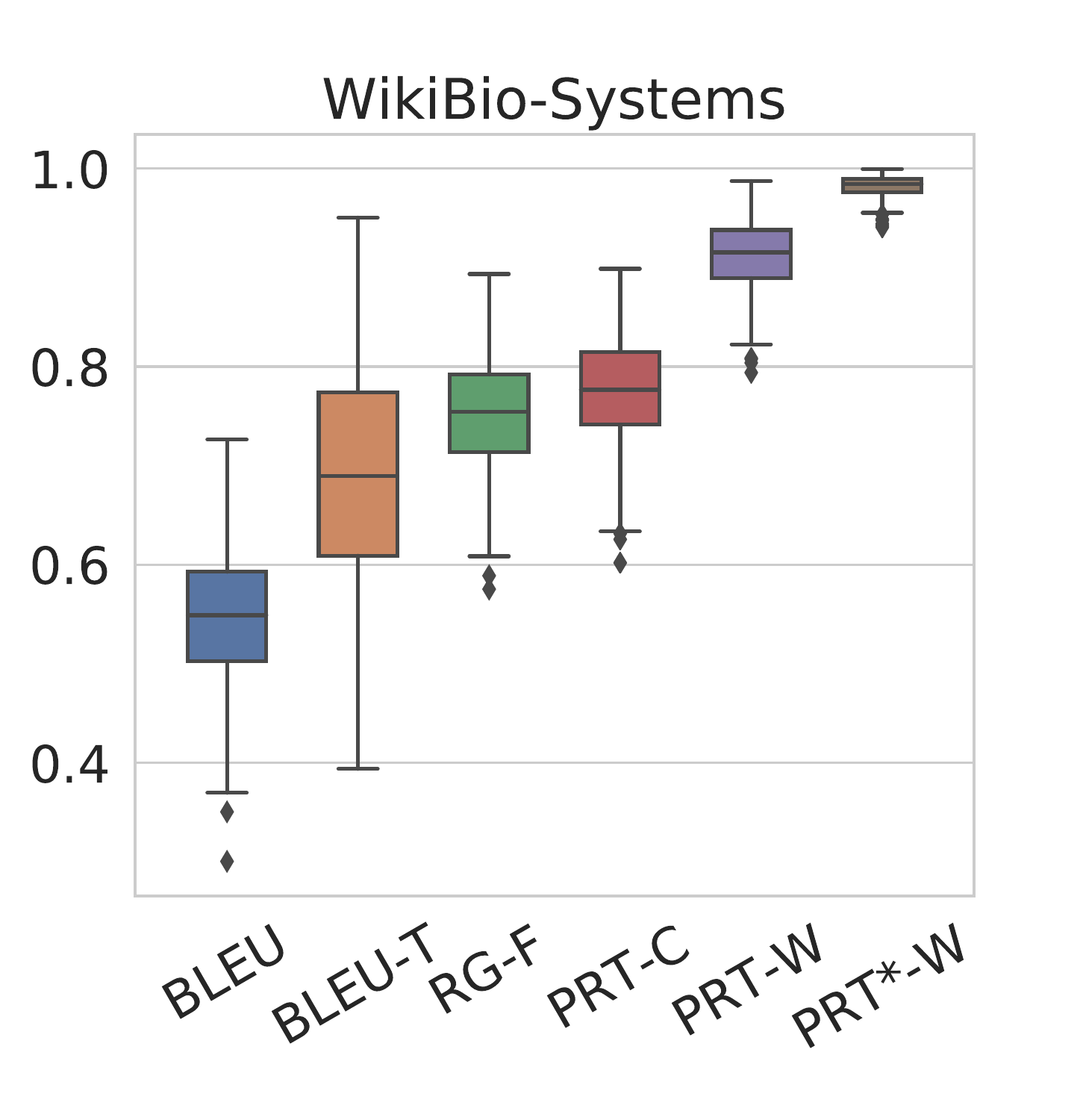}
        \includegraphics[width=0.49\linewidth]{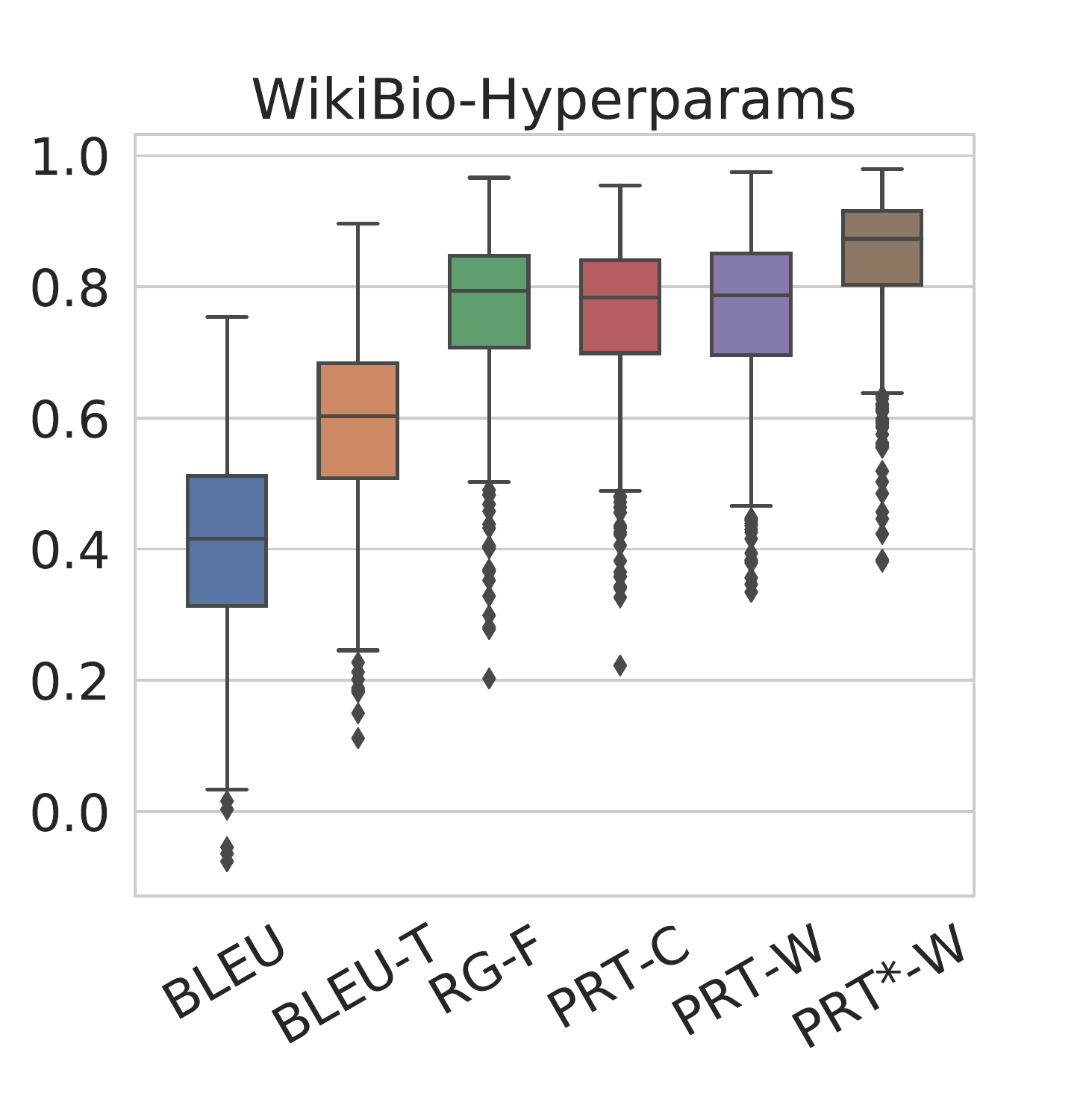}
    \caption{Distribution of metric correlations across $500$ bootstrap samples. PRT $=$ PARENT.}
\label{fig:correlations}
\end{figure}

We use bootstrap sampling ($500$ iterations)
over the $1100$ tables for which we
collected human annotations to get an idea of how the
correlation of each metric varies with the underlying data.
In each iteration, we sample with replacement, tables along with their references and all the generated texts
for that table. Then we compute aggregated human evaluation and metric scores for each of the models and compute the correlation between the two.
We report the average correlation across all bootstrap samples for each metric in Table~\ref{tab:results}.
The distribution of correlations for the best performing metrics are shown in Figure~\ref{fig:correlations}.

Table~\ref{tab:results} also indicates
whether PARENT is significantly better than a baseline metric.
\citet{graham2014testing} suggest using the William's test for this purpose, but since we are computing correlations between only $4$/$13$ systems at a time, this test has very weak power in our case. 
Hence, we use the bootstrap samples to obtain a $1-\alpha$
confidence interval of the difference in correlation between PARENT
and any other metric and check whether this is above $0$
\citep{wilcox2016comparing}.

Correlations are higher for the systems category than the hyperparams category.
The latter is a more difficult setting since very similar models are compared,
and hence the variance of the correlations is also high.
Commonly used metrics which only rely on the reference
(BLEU, ROUGE, METEOR, CIDEr)
have only weak correlations with human judgments.
In the hyperparams category, these are often negative, implying
that tuning models based on these may lead to
selecting worse models.
BLEU performs the best among these,
and adding n-grams from the table as references
improves this further (BLEU-T).

Among the extractive evaluation metrics, CS, which also only relies on the reference,
has poor correlation in the hyperparams category.
RG-F, and both variants of the PARENT metric achieve the highest correlation for
both settings. There is no significant difference among these for the hyperparams
category, but for systems,
PARENT-W is significantly better than the other two.
While RG-F needs a full information extraction pipeline in its implementation,
PARENT-C only relies on co-occurrence counts, and PARENT-W can be used
out-of-the-box for any dataset.
To our knowledge, this is the first rigorous evaluation
of using information extraction for generation evaluation.

On this dataset, the word-overlap model
showed higher correlation than the co-occurrence model
for entailment.
In \S \ref{sec:webnlg} we will show that for the WebNLG
dataset, where more paraphrasing is involved between the
table and text, the opposite is true.
Lastly, we note that the heuristic for selecting $\lambda$ is sufficient to produce high
correlations for PARENT, however,
if human annotations are available,
this can be tuned to produce significantly
higher correlations (\textit{PARENT*-W/C}).

\subsection{Analysis}
\label{sec:noise}

\begin{figure}
    \centering
        \includegraphics[width=0.49\linewidth]{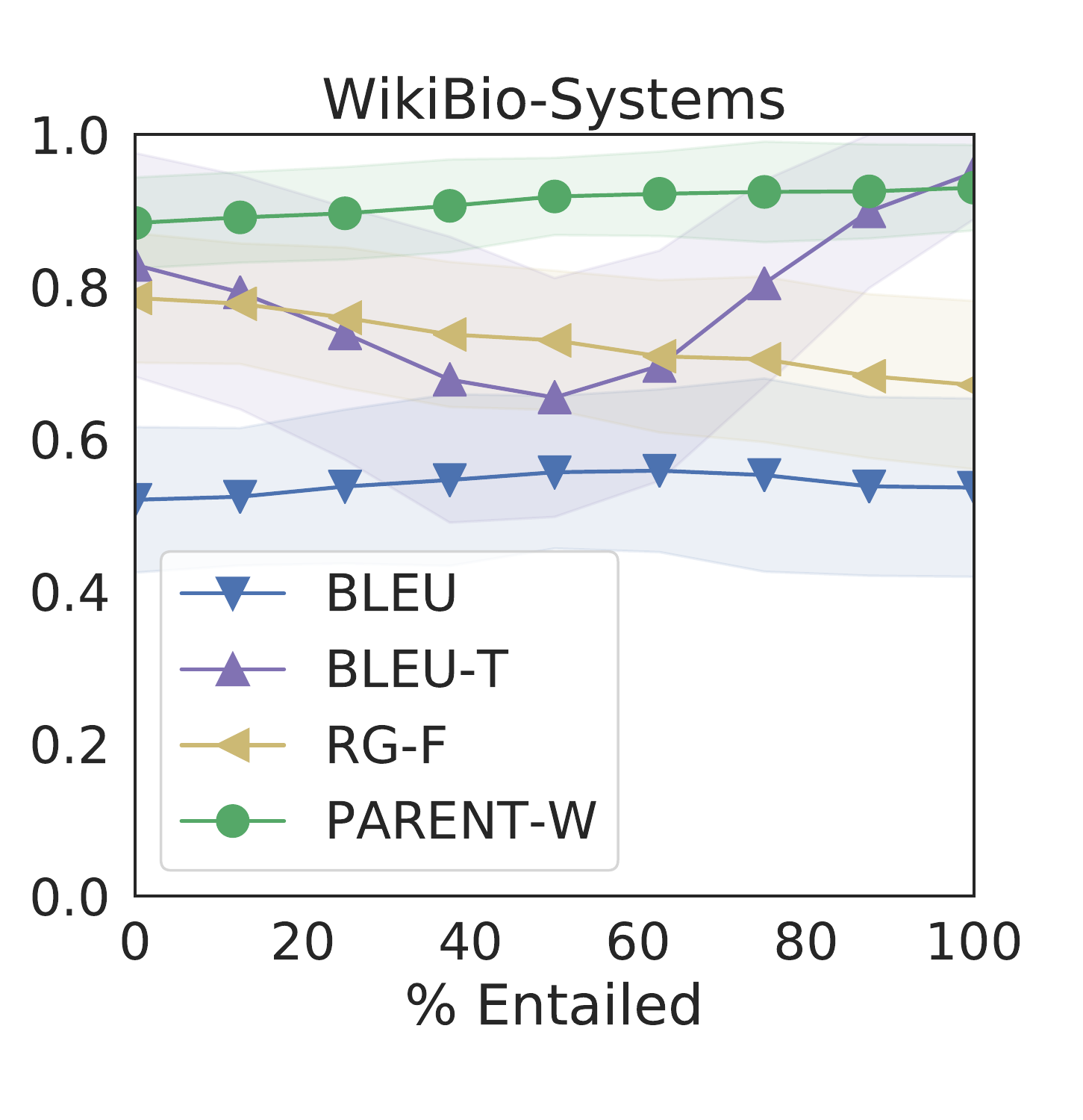}~
        \includegraphics[width=0.49\linewidth]{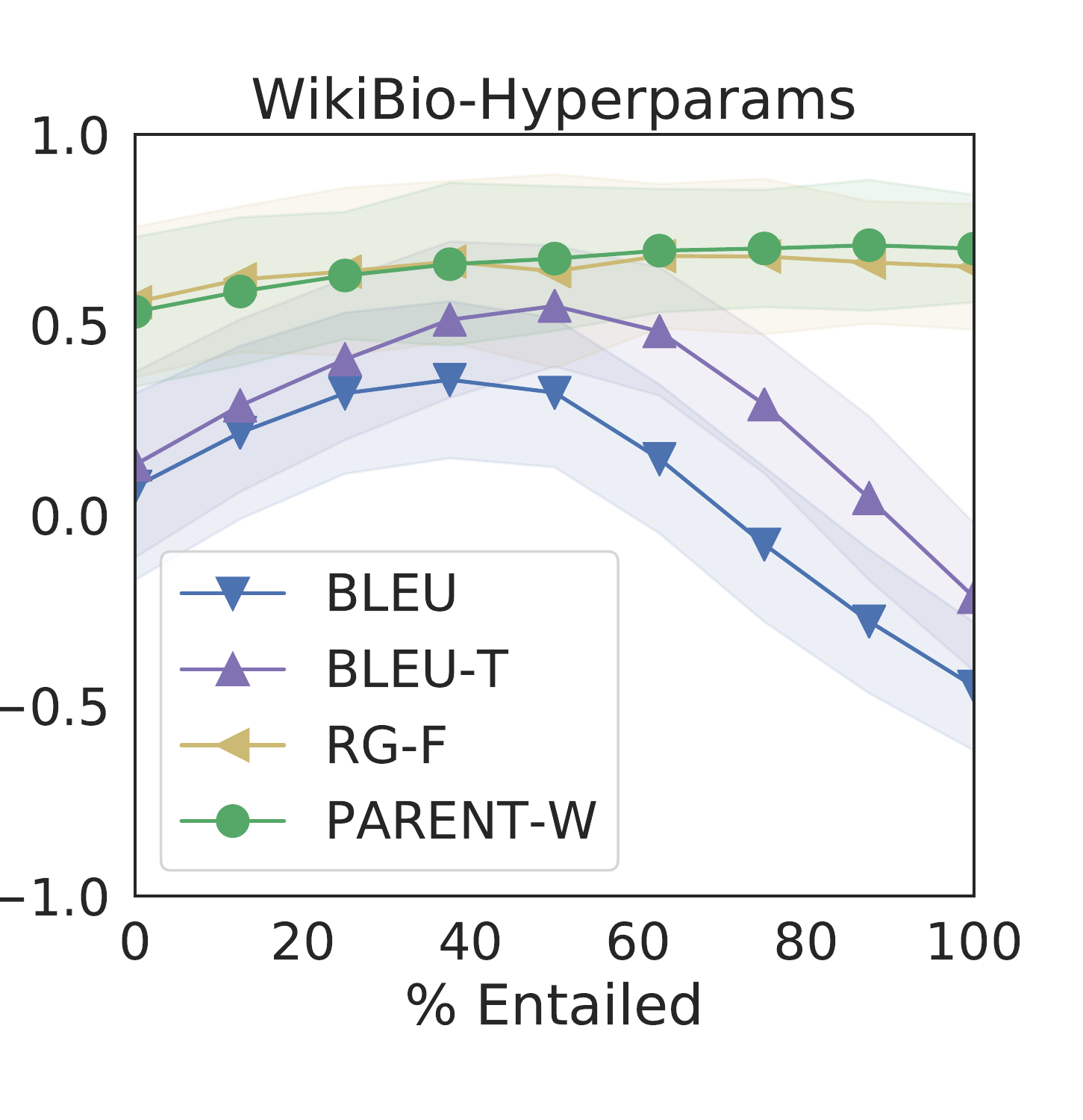}
    \caption{Correlation of the metrics to human judgment as the percentage of entailed examples in WikiBio is varied.}
    \label{fig:divergence}
\end{figure}

In this section we further analyze the performance of PARENT-W%
\footnote{The trends were similar for PARENT-C.}
under different conditions, and compare to the other best
metrics from Table~\ref{tab:results}.

\paragraph{Effect of Divergence.} To study the correlation as we vary the number of divergent references, we also collected binary labels from workers for whether a reference
is \textit{entailed} by the corresponding table. We define a reference as entailed when it mentions only information which can be inferred from the table. Each table and reference pair was judged by $3$ independent workers, and we used the majority vote as the label for that pair. Overall, only $38\%$ of the references were labeled as entailed by the table. Fleiss' $\kappa$  was $0.30$, which indicates a fair agreement. We found the workers sometimes disagreed on what information can be reasonably entailed by the table.

Figure~\ref{fig:divergence} shows the correlations
as we vary the percent of entailed examples in the evaluation set of WikiBio. 
Each point is obtained by fixing the desired proportion of entailed examples, and sampling subsets from the full set which satisfy this proportion. 
PARENT and RG-F remain stable and show a high correlation across the entire range,
whereas BLEU and BLEU-T vary a lot.
In the hyperparams category, the latter two have the
worst correlation when the evaluation set contains \textit{only entailed} examples, which may seem surprising.
However, on closer examination we found that this subset tends to omit a lot of information from the tables. Systems which produce more information than these references are penalized by BLEU, but not in the human evaluation.
PARENT overcomes this issue by measuring recall against the table in
addition to the reference.

\paragraph{Ablation Study.} We check how different components in the computation of PARENT contribute to its correlation to human judgments. Specifically, we remove the probability $w(g)$ of an n-gram $g$ being entailed by the table from Eqs.~\ref{eq:precision} and
\ref{eq:recall}.%
\footnote{When computing precision we set $w(g)=0$, and when computing recall we set $w(g)=1$ for all $g$.}
The average correlation for PARENT-W drops to $0.168$ in this case.
We also try a variant of PARENT with $\lambda=0$, which removes the contribution of Table Recall (Eq.~\ref{eq:lambda}).
The average correlation is $0.328$ in this case.
With these components, the correlation is $0.838$,
showing that they are crucial to the performance of PARENT.

\paragraph{Sentence Level Discrimination.} \citet{chaganty} point out that hill-climbing on an automatic metric is meaningless if that metric has a low \textit{instance-level} correlation to human judgments. In Table~\ref{tab:accuracy} we show the average accuracy of the metrics in making the same judgments as humans between pairs of generated texts.
Both variants of PARENT are significantly better than the other metrics,
however the best accuracy is only $60\%$ for the binary task.
This is a challenging task, since there are typically only
subtle differences between the texts.
Achieving higher instance-level accuracies will 
require more sophisticated language understanding models for evaluation.

\begin{table}[!tb]
\centering
\small
\begin{tabular}{@{}ccccc@{}}
\toprule
\textbf{BLEU} & \textbf{BLEU-T} & \textbf{RG-F} & \textbf{PARENT-W} & \textbf{PARENT-C} \\ \midrule
0.556         & 0.567$^\ast$           & 0.588$^\ast$       & 0.598$^\ddagger$          & \textbf{0.606}$^\dagger$          \\ \bottomrule
\end{tabular}
\caption{Accuracy on making the same judgments as humans between pairs of generated texts.
         $p<0.01^\ast/0.05^\dagger/0.10^\ddagger$: accuracy is
         significantly higher than the next best accuracy to the left using a paired McNemar's test.}
 \label{tab:accuracy}
\end{table}


\subsection{WebNLG Dataset}
\label{sec:webnlg}
\setlength{\tabcolsep}{3pt}

\begin{table}[!tbp]
\centering
\small
\begin{tabular}{@{}lccccccc@{}}
\toprule
\textbf{Metric} & \textbf{Grammar} & \textbf{Fluency} & \textbf{Semantics} & \textbf{Avg} \\ \midrule
METEOR          & 0.788$\pm$0.04     & 0.792$\pm$0.04     & 0.576$\pm$0.06     & 0.719            \\
ROUGE           & 0.788$\pm$0.04     & 0.792$\pm$0.04     & 0.576$\pm$0.06     & 0.719            \\
CIDEr           & 0.804$\pm$0.03     & 0.753$\pm$0.04     & 0.860$\pm$0.02     & 0.806            \\
BLEU            & \textbf{0.858$\pm$0.02}     & \textbf{0.811$\pm$0.03}     & 0.775$\pm$0.03     & 0.815            \\
BLEU-T            & 0.849$\pm$0.02     & 0.801$\pm$0.03     & 0.816$\pm$0.02     & 0.822            \\
CIDErD          & 0.838$\pm$0.04     & 0.796$\pm$0.04     & 0.853$\pm$0.02     & 0.829            \\ \midrule
PARENT-W           & 0.821$\pm$0.03     & 0.768$\pm$0.04     & \textbf{0.887$\pm$0.02}     & 0.825            \\
PARENT-C           & 0.851$\pm$0.03     & 0.809$\pm$0.04     & 0.877$\pm$0.02     & \textbf{0.846}            \\ \bottomrule
\end{tabular}
\caption{Average pearson correlation across $500$ bootstrap samples
        of each metric to human ratings for
        each aspect of the
        generations from the WebNLG challenge.} 
\label{tab:webnlg}
\end{table}

To check how PARENT correlates with human judgments when the references
are elicited from humans (and less likely to be divergent),
we check
its correlation with the human ratings provided for the systems
competing in the WebNLG challenge \citep{gardent2017creating}. 
The task is to generate text describing $1$-$5$ RDF triples
(e.g. \textit{John E Blaha, birthPlace, San Antonio}),
and human ratings were collected for the outputs of
$9$ participating systems on $223$ instances.
These systems include a mix of pipelined, statistical and neural methods.
Each instance has upto $3$ reference texts associated with the RDF
triples, which we use for evaluation.

The human ratings were collected on $3$ distinct aspects -- \textit{grammaticality},
\textit{fluency} and \textit{semantics}, where \textit{semantics} 
corresponds to the degree to which a generated text agrees
with the meaning of the underlying RDF triples.
We report the correlation of several metrics with these ratings in
Table~\ref{tab:webnlg}.\footnote{
We omit extractive evaluation metrics since no extraction
systems are publicly available for this dataset, and
developing one is beyond the scope of this work.}
Both variants of PARENT are either competitive or better
than the other metrics in terms of the average correlation
to all three aspects.
This shows that PARENT is applicable for high quality references
as well.

While BLEU has the highest correlation for the grammar and fluency
aspects,
PARENT
does best for semantics.
This suggests that the inclusion of source tables into the
evaluation orients the metric more towards measuring
the fidelity of the content of the generation.
A similar trend is seen comparing BLEU and BLEU-T.
As modern neural text generation systems are typically
very fluent, measuring their fidelity is of
increasing importance.
Between the two entailment models, PARENT-C is better
due to its
higher correlation with the grammaticality and fluency aspects.

\begin{figure}
    \centering
    \includegraphics[width=0.85\linewidth, trim={0.7cm 0 0 0 }, clip]{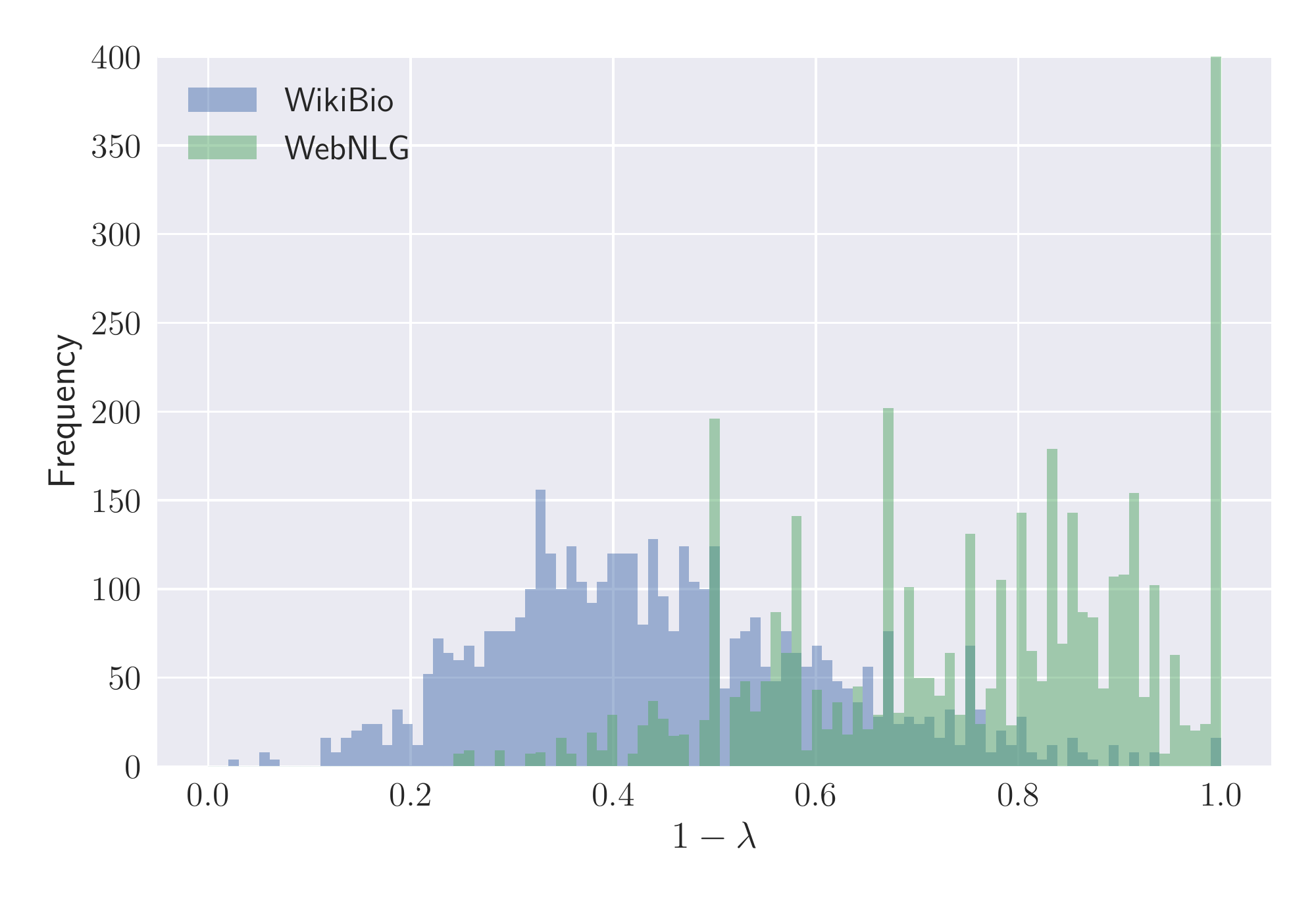}
    \caption{Histogram of the recall of the references against
    the table (Eq.~\ref{eq:table}), which is used to set
    $1-\lambda$.
    Lower values indicate that the metric
    relies more on the table and less on the reference.}
    \label{fig:lambda-hist}
\end{figure}

\paragraph{Distribution of $\lambda$.}
The $\lambda$ parameter in the calculation of PARENT
decides whether to compute recall against the table or
the reference (Eq.~\ref{eq:lambda}).
Figure~\ref{fig:lambda-hist} shows the distribution of
the values taken by $1-\lambda$ using the heuristic
described in \S \ref{sec:parent}
for instances in both WikiBio and WebNLG.
For WikiBio, the recall of the references against the
table is generally low, and hence the recall of the
generated text relies more on the table.
For WebNLG, where the references are elicited from
humans, this recall is much higher (often $1.0$),
and hence the recall of the generated text relies more
on the reference.

\section{Related Work}

Over the years several studies have evaluated automatic metrics for measuring text generation performance \citep{callison2006re,stent2005evaluating,belz2006comparing,reiter2018structured,liu2016not,kilickaya2017re,gatt2018survey}. The only consensus from these studies seems to be that no single metric is suitable across all tasks. A recurring theme is that metrics like BLEU and NIST \citep{doddington} are not suitable for judging content quality in NLG. Recently, \citet{novikova2017we} did a comprehensive study of several metrics on the outputs of state-of-the-art NLG systems, and found that while they showed acceptable correlation with human judgments at the \textit{system} level, they failed to show any correlation at the \textit{sentence} level.
Ours is the first study which checks the quality of metrics when table-to-text references are divergent. We show that in this case even system level correlations can be unreliable.

Hallucination \citep{rohrbach2018object,lee2018hallucinations} refers to when an NLG system generates
text which mentions extra information than what is present in the source
from which it is generated.
Divergence can be viewed as hallucination in the reference text itself.
PARENT deals with hallucination by discounting n-grams which do not
overlap with either the reference or the table.

PARENT draws inspiration from iBLEU \citep{sun2012joint}, a metric for evaluating paraphrase generation, which compares the generated text to both the source text and the reference. While iBLEU penalizes texts which match the source, here we reward such texts since our task values accuracy of generated text more than the need for paraphrasing the tabular content \cite{liu-dahlmeier-ng:2010:EMNLP}. Similar to SARI for text simplification \citep{xu2016optimizing} and Q-BLEU for question generation \citep{nema2018towards}, PARENT falls under the category of task-specific metrics.

\section{Conclusions}



We study the automatic evaluation of table-to-text systems when the references diverge from the table. 
We propose a new metric, PARENT, which shows the highest correlation
with humans across a range of settings with
divergent references in WikiBio.
We also perform the first empirical evaluation of
information extraction based metrics \citep{wiseman2017challenges},
and find RG-F to be effective.
Lastly, we show that PARENT is comparable
to the best existing metrics when references are elicited
by humans on the WebNLG data.


\section*{Acknowledgements}
Bhuwan Dhingra is supported by a fellowship from Siemens,
and by grants from Google.
We thank Maruan Al-Shedivat, Ian Tenney, Tom Kwiatkowski, Michael Collins,
Slav Petrov, Jason Baldridge, David Reitter and other members of the
Google AI Language team for helpful discussions and suggestions.
We thank Sam Wiseman for sharing data for an earlier version of
this paper.
We also thank the anonymous reviewers for their feedback.

\bibliography{tacl2018}
\bibliographystyle{acl_natbib}

\FloatBarrier

\appendix

\section{Appendices}

\subsection{Information Extraction System}
\label{app:ie-system}

For evaluation via information extraction \citep{wiseman2017challenges}
we train a model for WikiBio which accepts text as input
and generates a table as the output.
Tables in WikiBio are open-domain, without any fixed schema
for which attributes may be present or absent in an instance.
Hence we employ the Pointer-Generator Network (PG-Net) \citep{see2017pointer}
for this purpose.
Specifically, we use a sequence-to-sequence model, whose encoder
and decoder are both single-layer bi-directional LSTMs.
The decoder is augmented with an attention mechanism over
the states of the encoder. 
Further, it also uses a copy mechanism
to optionally copy tokens directly from the source text.
We do not use the coverage mechanism of \citet{see2017pointer}
since that is specific to the task of summarization they study.
The decoder is trained to produce a linearized version of the table
where the rows and columns are flattened into a sequence,
and separate by special tokens.
Figure~\ref{fig:text-to-table} shows an example.

\begin{figure}
    \centering
    \small
    \begin{tabular}{l}
        \hline
        \textbf{Text:}\\
        michael dahlquist ( december 22 , 1965 -- july 14 , 2005 )\\
        was a drummer in the seattle band silkworm .\\
        \hline
        \textbf{Table:}\\
        name $<$C$>$ michael dahlquist $<$R$>$ birth date $<$C$>$ 22\\
        december 1965 $<$R$>$ birth place $<$C$>$ seattle , washington\\
        $<$R$>$ death date $<$C$>$ 14 july 2005 $<$R$>$ death\\
        place $<$C$>$ skokie , illinois $<$R$>$ genres $<$C$>$ male\\
        $<$R$>$ occupation(s) $<$C$>$ drummer $<$R$>$ instrument\\
        $<$C$>$ drums\\
        \hline
    \end{tabular}
    \caption{An input-output pair for the information extraction system.
    $<$R$>$ and $<$C$>$ are special symbols used to separate (attribute, value)
    pairs and attributes from values, respectively.}
    \label{fig:text-to-table}
\end{figure}

Clearly, since the references are divergent,
the model cannot be expected to produce the entire table,
and we see some false information being hallucinated after
training. Nevertheless, as we show in \S \ref{sec:results}, this
system can be used for evaluating generated texts.
After training, we can parse the output sequence along
the special tokens $<$R$>$ and $<$C$>$ to get a set
of (attribute, value) pairs.
Table~\ref{tab:ie-results} shows the precision, recall
and F-score of these extracted pairs against the ground
truth tables, where the attributes and values are
compared using an exact string match.

\begin{table}[]
\centering
\small
\begin{tabular}{@{}ccc@{}}
\toprule
\textbf{Precision} & \textbf{Recall} & \textbf{F-score} \\ \midrule
0.428              & 0.310           & 0.351            \\ \bottomrule
\end{tabular}
\caption{Performance of the Information Extraction system.}
\label{tab:ie-results}
\end{table}

\begin{table*}[]
\scriptsize
\centering
\begin{tabular}{@{}ll@{}}
\toprule
\textbf{Reference}  & vedran nikšić ( born 5 may 1987 in osijek ) is a croatian football striker . {[}STOP{]}                                                                                                                                             \\
\textbf{Prediction} & vedran nikšić ( born 5 may 1987 ) is a croatian football forward \textbf{who is currently a free agent} . {[}STOP{]}                                                                                                                         \\ \midrule
\textbf{Reference}  & \begin{tabular}[c]{@{}l@{}}adam whitehead ( born 28 march 1980 ) is a former breaststroke swimmer from coventry , england , \textit{who competed at the 2000}\\ \textit{summer olympics in sydney , australia} . {[}STOP{]}\end{tabular}              \\
\textbf{Prediction} & adam whitehead ( born 28 march 1980 ) is an english swimmer . {[}STOP{]}                                                                                                                                                            \\ \midrule
\textbf{Reference}  & chris fortier is an american dj and founder of the balance record pool \textit{as well as co-founder and owner of fade records} . {[}STOP{]}                                                                                                 \\
\textbf{Prediction} & chris fortier ( born in melbourne , florida ) is an american disc jockey and record producer \textit{from melbourne , florida} . {[}STOP{]}                                                                                                  \\ \midrule
\textbf{Reference}  & pretty balanced was an american band based in columbus , ohio . {[}STOP{]}                                                                                                                                                          \\
\textbf{Prediction} & pretty balanced is an american piano band from columbus , ohio . {[}STOP{]}                                                                                                                                                         \\ \midrule
\textbf{Reference}  & \begin{tabular}[c]{@{}l@{}}ben street ( born february 13 , 1987 ) is a canadian professional ice hockey player who is a member within the colorado avalanche\\ organization of the national hockey league . {[}STOP{]}\end{tabular} \\
\textbf{Prediction} & \begin{tabular}[c]{@{}l@{}}ben street ( born february 13 , 1987 ) is a canadian professional ice hockey centre currently playing for the colorado avalanche of\\ the national hockey league ( nhl ) . {[}STOP{]}\end{tabular}       \\ \bottomrule
\end{tabular}
\caption{Sample references and predictions from PG-Net with beam size $8$.
Information which is absent from the reference, but can be inferred from the table is
in \textbf{bold}.
Information which is present in the reference, but cannot be inferred from the table is
in \textit{italics}.}
\label{tab:examples}
\end{table*}

\subsection{Hyperparameters}
\label{app:hyperparams}

After tuning we found the same set of hyperparameters to work well for
both the table-to-text PG-Net, and the inverse information extraction
PG-Net.
The hidden state size of the biLSTMs was set to $200$.
The input and output vocabularies were set to $50000$ most common
words in the corpus, with additional special symbols for table attribute
names (such as ``birth-date''). The embeddings of the tokens
in the vocabulary were initialized with Glove \citep{pennington2014glove}.
Learning rate of $0.0003$ was used during training, with the Adam
optimizer, and a dropout of $0.2$ was also applied to the outputs
of the biLSTM.
Models were trained till the loss on the dev set stopped dropping.
Maximum length of a decoded text was set to $40$ tokens, and that
of the tables was set to $120$ tokens.
Various beam sizes and length normalization penalties were applied 
for the table-to-text system, which are listed in the main paper.
For the information extraction system, we found a beam size of $8$
and no length penalty to produce the highest F-score on the dev set.

\subsection{Sample Outputs}
\label{app:outputs}

Table~\ref{tab:examples} shows some sample references and the corresponding predictions
from the best performing model, PG-Net for WikiBio.

\end{document}